\documentclass{article}

\usepackage{microtype}
\usepackage{graphicx}
\usepackage{booktabs} 

 \usepackage[accepted]{icml2023}

\usepackage{amsmath}
\usepackage{amssymb}
\usepackage{mathtools}
\usepackage{amsthm}

\theoremstyle{plain}
\newtheorem{theorem}{Theorem}[section]

\theoremstyle{definition}
\newtheorem{definition}[theorem]{Definition}

\theoremstyle{remark}

\usepackage{algorithm}
\usepackage{algorithmic}

\usepackage{nicefrac}
\usepackage{comment}
\usepackage{multirow}

\icmltitlerunning{Learning to Play Stochastic Two-player Perfect-Information Games without Knowledge}

\begin{document}

\twocolumn[
\icmltitle{Learning to Play Stochastic Two-player Perfect-Information Games \\ without Knowledge}



\icmlsetsymbol{equal}{*}

\begin{icmlauthorlist}
\icmlauthor{Quentin Cohen-Solal}{affiliation}
\icmlauthor{Tristan	Cazenave}{affiliation}

\end{icmlauthorlist}

\icmlaffiliation{affiliation}{LAMSADE, Université Paris-Dauphine, PSL, CNRS}

\icmlcorrespondingauthor{Quentin Cohen-Solal}{quentin.cohen-solal@dauphine.psl.eu}
\icmlcorrespondingauthor{Tristan Cazenave}{cazenave@lamsade.dauphine.fr}

\icmlkeywords{Reinforcement Learning, Stochastic Games, Tree Search}

\vskip 0.3in
]



\printAffiliationsAndNotice{}  

\begin{abstract}

In this paper, we extend the Descent framework, which enables learning and planning in the context of two-player games with perfect information, to the framework of stochastic games.

We propose two ways of doing this, the first way generalizes the search algorithm, i.e. Descent, to stochastic games and the second way approximates stochastic games by deterministic games.

 We then evaluate them on the game \emph{EinStein würfelt nicht!} against state-of-the-art algorithms: \emph{Expectiminimax} and Polygames (i.e. the \emph{Alpha Zero} algorithm). It is our generalization of Descent which obtains the best results. The approximation by deterministic games nevertheless obtains good results, presaging that it could give better results in particular contexts.

\end{abstract}

\noindent\begin{minipage}[t]{1\columnwidth}%
\global\long\def\et{\ \wedge\ }%

\global\long\def\terminal{\mathrm{t}}%

\global\long\def\joueur{\mathrm{j}}%

\global\long\def\joueurUn{\mathrm{1}}%

\global\long\def\joueurDeux{\mathrm{2}}%

\global\long\def\fbin{\mathrm{f_{b}}}%

\global\long\def\Actions{\mathcal{A}}%

\global\long\def\actions#1{\mathrm{actions}\left(#1\right)}%

\global\long\def\etats{\mathcal{S}}%
\end{minipage}

\global\long\def\terminalrandom{\mathrm{t_{r}}}%
\global\long\def\hfb{\mathrm{b_{t}}}%
\global\long\def\hfp{\mathrm{p_{t}}}%
\global\long\def\hadapt{f_{\theta}}%
\global\long\def\hadaptnum#1{f_{\theta_{#1}}}%
\global\long\def\itreeset{\mathbf{T_{i}}}%
\global\long\def\treeset{\mathbf{T}}%
\global\long\def\rootset{\mathbf{R}}%
\global\long\def\hterminal{f_{\mathrm{t}}}%
\global\long\def\rnap{\mathrm{p_{rna}}}%
\global\long\def\rnab{\mathrm{b_{rna}}}%
\global\long\def\ubfm{\mathrm{UBFM}}%
\global\long\def\ubfmt{\ubfm_{\mathrm{s}}}%
\global\long\def\argmax{\operatorname*{\mathrm{arg\,max}}}%
\global\long\def\argmin{\operatorname*{\mathrm{arg\,min}}}%
\global\long\def\liste#1#2{\left\{  #1\,|\,#2\right\}  }%
\global\long\def\fterminal{\hterminal}%
\global\long\def\fadapt{\hadapt}%
\global\long\def\id{\mathrm{ID}}%
\global\long\def\minimum{\operatorname*{\mathrm{min}}}%

\global\long\def\initial{\mathrm{i}}%
\global\long\def\vide{\varnothing}%

\global\long\def\Spartiel{\mathcal{S}_{\mathrm{p}}}%

\global\long\def\et{\ \wedge\ }%
\global\long\def\terminal{\mathrm{t}}%
\global\long\def\joueur{\mathrm{j}}%
\global\long\def\joueurUn{\mathrm{1}}%
\global\long\def\joueurDeux{\mathrm{2}}%
\global\long\def\fbin{\mathrm{f_{b}}}%
\global\long\def\fterminal{f_{\mathrm{t}}}%
\global\long\def\fadapt{f_{\theta}}%
\global\long\def\actions{\mathcal{A}}%
\global\long\def\algo{\mathscr{A}}%
\global\long\def\continue{\mathrm{continue}}%
\global\long\def\prioritychildcalculation{\mathrm{calculate\_priority\_child}}%

\global\long\def\etats{\mathcal{S}}%
\global\long\def\Spartiel{\mathcal{S}_{\mathrm{p}}}%
\global\long\def\ubfm{\mathrm{UBFM}}%
\global\long\def\ubfms{\ubfm_{\mathrm{\mathrm{s}}}}%
\global\long\def\umaxn{\ubfm^{n}}%
\global\long\def\maxn{\mathrm{Max}^{n}}%
\global\long\def\umaxns{\ubfms^{n}}%
\global\long\def\descente{\mathrm{Descent}}%
\global\long\def\descenten{\descente^{n}}%
\global\long\def\fbinn{\fbin}%
\global\long\def\fterminaln{\fterminal}%
\global\long\def\fadaptn{\fadapt}%
\global\long\def\ou{\,\vee\,}%
\global\long\def\ubfm{\mathrm{UBFM}}%
\global\long\def\argmax{\operatorname*{\mathrm{arg\,max}}}%
\global\long\def\argmin{\operatorname*{\mathrm{arg\,min}}}%
\global\long\def\liste#1#2{\left\{  #1\,|\,#2\right\}  }%
\global\long\def\minimum{\operatorname*{\mathrm{min}}}%
\global\long\def\initial{\mathrm{i}}%
\global\long\def\vide{\varnothing}%

\section{Introduction}

Deep reinforcement learning algorithms enable to surpass human players in many two-player complete information games such as chess and Go \cite{silver2018general,cazenave2020polygames,cohen2020learning,cohen2021minimax}. General frameworks for learning to play perfect information games such as Polygames \cite{cazenave2020polygames} have also been applied to stochastic two-player perfect-information games such as \emph{EinStein würfelt nicht!}. They use an \emph{AlphaZero} like deep reinforcement learning algorithm to handle stochastic games where they just add random nodes to usual minimization and maximization nodes. The Descent framework \cite{cohen2020learning,cohen2021minimax} is an alternative deep reinforcement learning algorithm that learns by self play starting with zero knowledge. It is quite different from \emph{AlphaZero}  frameworks since it does not use MCTS but instead two search algorithms: \emph{Unbounded Minimax} and \emph{Descent Minimax}. In addition, it does not use a policy but only a value network. In this paper we extend the Descent framework for dealing with stochastic two-player perfect-information games.

We start by  
detailing
the algorithms of the literature used in this article,
namely \emph{AlphaZero}, \emph{Descent Minimax}, and \emph{Expectiminimax} (Section~\ref{subsec:Literature-algorithms}) and the game EinStein würfelt nicht! (Section \ref{games}).
Then, we present our contributions (Section~\ref{subsec:New-Algorithms})
.
In particular, we introduce
a technique allowing to apply \emph{Descent Minimax} without modifying it in the
context of stochastic games: by approximating the stochastic game
with a set of deterministic games (Section~\ref{subsec:Approximate-a-Stochastic} and \ref{formalization_determinization}).
We also introduce a generalization of \emph{Descent Minimax} in order to directly
learn to play stochastic games without having to perform approximations
(Section~\ref{subsec:Descent-for-Stochastic}). Moreover, in Section~\ref{subsec:Comparison-Experiments},
we compare our two contributions with each other 
and with the main state-of-the-art algorithms: \emph{Expectiminimax}
and \emph{AlphaZero}
.

\section{Background}

\subsection{Used Literature Algorithms\label{subsec:Literature-algorithms}}

We now detail the literature game search algorithms we use.

\subsubsection{AlphaZero / Polygames}

We start by presenting the \emph{AlphaZero} algorithm \cite{silver2018general} and one of its implementations,
named Polygames \cite{cazenave2020polygames}, which we use in the experiments of this article.
\emph{AlphaZero} / Polygames uses its search algorithm, MCTS with PUCT, to generate matches,
by playing against itself. It uses the data from these matches
to update its neural network. This neural network is used by the search
algorithm to evaluate states by a value and by a policy (i.e. a probability
distribution on the actions playable in that state). For each finished
match, the network is trained to associate with each state of the
state sequence of this match the result of the end of that match (which
is $-1$ for a loss,  $0$ for a draw, and $+ 1$ for a win). It is also trained, at the same time, to associate
with each state a particular policy. In that ``target'' policy, the probability of an action
 is proportional to $N^{\nicefrac{1}{\tau}}$
where $N$ is the number of times this action has been selected in
the search from that state and $\tau$ is a parameter.

\subsubsection{Descent Minimax}

We now present the framework of the Descent algorithm \cite{cohen2020learning}.
The learning framework of Descent is based on \emph{Unbounded Minimax} \cite{korf1996best}: an algorithm calculating an approximation of the minimax value of a game state ;  and on 
 \emph{Descent Minimax}: a variant of Unbounded Minimax
which consists in exploring the sequences of actions until terminal
states. In comparison, Unbounded Minimax and MCTS explore a sequence
of actions only until a leaf state is reached. An iteration of Descent
thus consists in a deterministic complete simulation of the rest of
the game. The exploration is thus deeper while remaining a best-first
approach. This allows the values of terminal states to be propagated
more quickly to (shallower) non-terminal states.

Unlike Polygames, the learned target value of a state is not the end-game
value but its minimax value of the partial game tree built during
the match. This information is more informative, since it contains
part of the knowledge acquired during the previous matches. In addition,
contrary to Polygames, learning is carried out for each state of the
partial game tree constructed during the searches of the match (not
just for each state of the states sequence of the played match). In
other words, with Polygames, there is one learning target per search
whereas with the Descent framework, there are several learning targets
per search. Therefore, there is no
loss of information with Descent: all of the information acquired during the search
is used during the learning process. As a result, the Descent framework
generates a much larger amount of data for training from the same
number of played matches than AlphaZero / Polygames. Thus, unlike
the state of the art which requires to generate matches in parallel
to build its learning dataset, this approach does not uses the
parallelization of matches.

Finally, this approach is optionally based on a \emph{reinforcement
heuristic}, that is to say an evaluation function of terminal states
more expressive than the classical gain of a game (i.e. $+1$ / $0$
/ $-1$). The best proposed general reinforcement heuristics in \cite{cohen2020learning}
are \emph{scoring} and the \emph{depth heuristic} (the latter favoring
quick wins and slow defeats). 

Note that this approach does not use a policy, so there is no need
to encode actions. Consequently, this avoids the learning performance
problem of neural networks for games with large number of actions
(i.e. very large output size). In addition, although the Descent framework
does not perform matches in parallel, it batches all the child states
of an extended state together to be evaluated at one time on the GPU \cite{cohen2020learning}.

\subsubsection{Expectiminimax}

To end this section, we present the \emph{Expectiminimax} algorithm \cite{michie1966game,russell2009artificial}, which
is a generalization of the \emph{Minimax} algorithm in the context of stochastic
games. 
Recall that with \emph{Minimax}, the game trees have max state nodes (decision nodes of the first player where the latter maximizes the value of the next state node evaluated from his point of view) and min state nodes (decision nodes of the second player where this one minimizes the value of the next state node evaluated from the point of view of its adversary). With \emph{Expectiminimax}, there are also chance state nodes (nodes where "nature" plays). The value of a chance node is the weighted average of the values of its child state nodes by their probability of occurrence.

The version of  \emph{Expectiminimax} that we use, uses the following
improvement:  \emph{iterative deepening}  \cite{korf1985depth,slate1983chess}. With this improvement, \emph{Expectiminimax} uses a time limit. Several searches whose depth is each time increased are  performed sequentially on the same state as long as time remains. 

\subsection{Tested Games\label{games}}

We present in this section the game \emph{EinStein würfelt nicht!}, on which
we perform our algorithm comparison experiments for zero-knowledge
reinforcement learning.

\begin{figure}
\begin{centering}
\includegraphics[scale=0.15]{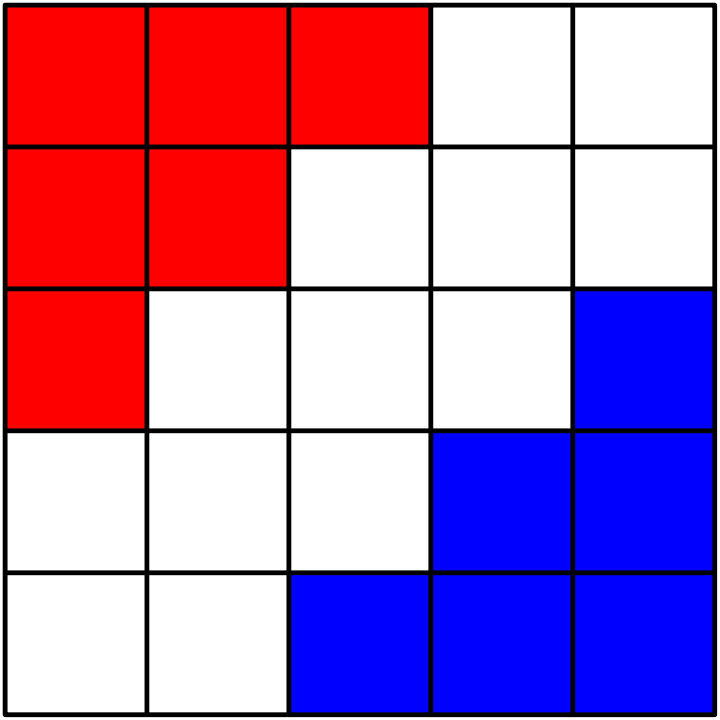}\caption{The two starting areas at EinStein wuerfelt nicht.\label{einstein}}
\par\end{centering}
\end{figure}

The game is played on a $5\times5$ square board. The objective is to reach with one of his pieces his arrival position which is the square of the corner of the board located in the opponent's starting area. The two starting areas are described in Figure \ref{einstein}. The game has an initialization phase where the players will place their 6 pieces on the board. Each player's pieces are numbered from 1 to 6.
During the initialization phase, the first player places his pieces in his starting area and then the second player places his pieces in his starting area symmetrically.
Then a race phase starts. As soon as a player reaches his arrival square with one of his pieces, the game is over. At the start of each turn, during the race phase, a 6-sided die is rolled. It indicates the current player's piece that he can move (the one with the same number as the result of the die). If there is no piece on the board corresponding to the number on the die, the piece to be moved is the one whose number is closest to the result of the die. If there are two possibilities, the player chooses. A piece's possible move is one of three moves to an adjacent square that bring the piece closer to that player's arrival square (horizontal move, vertical move, or diagonal move). If the piece ends its movement on another piece (no matter which player it belongs to), the other piece is removed from the board.

It should be noted that there are several variants with respect to the initialization phase, where for example the pieces are arranged randomly. In the Ludii \cite{Piette2020Ludii} implementation, during the initialization phase, in turn, each player places one of his pieces in his starting area. 

Despite its great simplicity and the element of luck, it is a game that has tactics and strategies.

The more recent programs at \emph{EinStein würfelt nicht!} are based on MCTS \cite{lorentz2011mcts,lu2015einstein,chu2017agent}. There are also works on endgame tablebase \cite{turner2012einstein} and on solving \emph{EinStein würfelt nicht!} \cite{bonnet2017toward}.

\section{Contributions\label{sec:Contributions}}

Now, we present our contributions, starting by introducing a technique
to apply \emph{Descent Minimax} in the context of stochastic games as well as a generalization
of \emph{Descent Minimax} to be applicable without having to approximate stochastic
games. Then, we compare these two techniques with each other and with
the algorithms of the literature.

\subsection{New Algorithms\label{subsec:New-Algorithms}}

\subsubsection{Approximate a Stochastic Game by a Set of Deterministic Games\label{subsec:Approximate-a-Stochastic}}

Thus, we start by introducing a technique to apply Descent (and any
other search algorithm for deterministic games) on a stochastic game.

This technique, which we call stochastic game determinization, approximates
a stochastic game by a set of deterministic games. This set corresponds
to all the possible realizations of the random events present in the
stochastic game. Each game of this set thus corresponds to a realization
of the random events of the stochastic game. This realization is known
to the players before the start of the game, the games of this set
are thus without chance. For example, in the case of a dice game,
this amounts on the one hand to rolling all the dice of the game in
advance and that the results of these rolls are known before the start
of the game. Players can then plan according to the results of the
dice, since they are known in advance.

In practice (for computational reasons), a random event is not resolved
before the start of the game but only when a player needs it, i.e.
when it occurs during the game or during the search made by one of
the players. Therefore, each time this event is encountered again
(during the same search, during the search of the other player, or
later in the same match), the result of the event will be the same.

The idea of this technique is to perform a reinforcement learning
process using this game set to learn how to play the stochastic game.
As with classic game learning techniques, a sequence of matches and
the respective learning phases are carried out as long as there is
time left. Moreover, on these matches, the chosen learning algorithm
for deterministic games is applied without modification. 

There are, however, two differences with the deterministic framework.
The first difference makes it possible to apply algorithms for deterministic
games to the stochastic game. It is the transformation of the stochastic
game into a set of deterministic games. More precisely, during the
reinforcement learning process, the games of the sequence of matches
are not the stochastic game but games of the determinization set (the
set of deterministic games corresponding to the stochastic game).
The games of the determinization set are selected randomly (according
to a certain probability corresponding to the probabilities of the
game).

The second difference with the deterministic framework is that there
is an additional step at the end of the match which transforms the
data obtained during the deterministic match into data for the stochastic
game. Data about chance states corresponding to the played deterministic
match are added to the data of the match. More precisely, the stochastic
match corresponding to the played deterministic match is implicitly
reconstructed and its chance states are added to be used during the
learning phases. The learning target value of these chance states
are estimated from this reconstruction. The learning phase will thus
make it possible to learn to play the stochastic game (but not necessarily
optimally). Then, during a confrontation (an evaluation match, a tournament,
etc.), the adaptive evaluation function learned during the reinforcement
learning process can be used in a conventional way, i.e. by a planning
algorithm for stochastic games, such as \emph{Expectiminimax}.

The advantage of this approach, compared to the one we are going to
present in the next section (and compared to other algorithms of the
same kind, using for example \emph{Expectiminimax}) is that the learning
is done on simplified games. Trees in these games are significantly
smaller and powerful 
pruning can now be performed. This
advantage should be particularly marked for games where the number
of possible outcomes of the random events is large. This probably
includes most card games. This probably does not include dice games
where one event is reduced to a single 6-sided dice roll or a coin
flipping (and where the number of such events is low over the course
of the game).

Indeed, for these games, with \emph{Expectiminimax} (or with generalized
Descent introduced in the next section), the number of children of
a chance node is the number of possible outcomes of the associated
random event. And, with these two algorithms, each outcome must be
explored, which is therefore costly if the number of outcomes is large.

The disadvantages of this approach are that it requires a game simulator
capable of performing the determinization and that the convergence
towards the optimal stochastic game strategy may not be guaranteed
in general.

\subsubsection{Formalization\label{formalization_determinization}}

We now formalize the concept of determinization of a stochastic game.
For this we recall the definition of deterministic two-player zero-sum
games of \cite{cohen2021completeness}. Then, we generalize it to
stochastic two-player zero-sum games. Finally, we formalize the process
of determinization for these stochastic games, which transforms such
a stochastic game into a set of deterministic games.

Thus, we start by recalling this definition of deterministic zero-sum
two-player games:
\begin{definition}
\cite{cohen2021completeness} A perfect two-player game is a tuple
$\left(S,\actions,\joueur,\fbin\right)$ where:
\begin{itemize}
\item $\left(\etats,\Actions\right)$ is a finite lower semilattice\footnote{i.e. $\left(\etats,\Actions\right)$ is a finite directed acyclic
graph satisfying that there exists $s_{0}\in\etats$ such that for
all $s\in\etats$, we have $s_{0}\leq s$ with $\leq$ the reachability
relation of $\left(\etats,\Actions\right)$.}, 
\item $\joueur$ is a function from $\etats$ to $\left\{ 1,2\right\} $, 
\item $\fbin$ is a function from $\liste{s\in\etats}{\terminal\left(s\right)}$
to $\left\{ -1,0,1\right\} $, where:
\begin{itemize}
\item $\terminal$ is a predicate such that $\terminal\left(s\right)$ is
true if and only if $\left|\Actions\left(s\right)\right|=0$ ; 
\item $\Actions\left(s\right)$ is the set defined by $\liste{s'\in\etats}{\left(s,s'\right)\in\Actions}$,
for all $s\in\etats$. 
\end{itemize}
\end{itemize}
\end{definition}

The set $\etats$ is the set of game states. $\Actions$ encodes the
actions of the game: $\Actions\left(s\right)$ is the set of states
reachable by an action from $s$. The function $\joueur$ indicates,
for each state, the number of the player whose turn it is, i.e. the
player who must play. The predicate $\terminal\left(s\right)$ indicates
if $s$ is a terminal state (i.e. an end-of-game state). Let $s\in\etats$
such that $\terminal\left(s\right)$ is true, the value $\fbin\left(s\right)$
is the payout for the first player in the terminal state $s$ (the
gain for the second player is $-\fbin\left(s\right)$ ; we are in
the context of zero-sum games). We have $\fbin\left(s\right)=1$ if
the first player wins, $\fbin\left(s\right)=0$ in the event of a
draw, and $\fbin\left(s\right)=-1$ if the first player loses.

We therefore generalize this definition to the context of stochastic
games using $S_{0}$ the set of chance nodes and $P$ describing the
probabilities of the different outcomes of chance nodes ($P\left(s,s'\right)$
is the probability to go from the state $s$ to the state $s'$):
\begin{definition}
A stochastic perfect two-player game is a tuple $\left(S,\Actions,\joueur,\fbin,P\right)$
where:
\begin{itemize}
\item $\left(\etats,\Actions\right)$ is a finite lower semilattice, 
\item $\joueur$ is a function from $\etats$ to $\left\{ 0,1,2\right\} $, 
\item $\fbin$ is a function from $\liste{s\in\etats}{\terminal\left(s\right)}$
to $\left\{ -1,0,1\right\} $, where:
\begin{itemize}
\item $\terminal$ is a predicate such that $\terminal\left(s\right)$ is
true if and only if $\left|\Actions\left(s\right)\right|=0$ ; 
\item $\Actions\left(s\right)$ is the set defined by $\liste{s'\in\etats}{\left(s,s'\right)\in\Actions}$,
for all $s\in\etats$,
\end{itemize}
\item $P$ is the \emph{game probabilities}, a function from $\etats_{0}\times\etats$
to $[0,1]$, with $\etats_{0}=\liste{s\in\etats}{\joueur\left(s\right)=0}$,
satisfying:
\begin{itemize}
\item for all $s\in\etats_{0}$, $\sum_{s'\in\Actions\left(s\right)}P\left(s,s'\right)=1$
,
\item for all $s\in\etats_{0}$, for all $s'\in\etats$ such that $\left(s,s'\right)\notin\Actions$,
$P\left(s,s'\right)=0$.
\end{itemize}
\end{itemize}
\end{definition}
We now assume that stochastic games are \emph{stochastically normalized},
i.e. for all $s\in\etats_{0}=\liste{s\in\etats}{\joueur\left(s\right)=0}$, for all $s'\in \Actions\left(s\right)$, 
$\joueur\left(s'\right)\neq0$. This assumption does not restrict
the applicability of the determinization technique in terms of games,
since any stochastic game can be reformulated into a stochastically
normalized stochastic game. The determinization could also be applied
without the normalized assumption, but the formalization would be
much more complicated.

We now formalize the process of determinization of a stochastic game,
which associate any stochastically normalized stochastic game $G$
with a deterministic game set $D(G)$. It is based on $S_{1,2}$ the
set of player nodes, $\delta\left(G\right)$ the set of functions
$d$ that associates each chance node with one of its children (the
determinization), $s^{\initial}$ the initial state of $G$, $s_{d}^{\initial}$
the initial state of the determinized game associated to $d$.
\begin{definition}
Let $G=\left(S,\Actions,\joueur,\fbin,P\right)$ be a stochastically
normalized stochastic perfect two-player game. The \emph{determinization}
of $G$ is the following set $D(G)=\liste{\left(S_{d},\Actions_{d},\joueur,\fbin\right)}{d\in\delta\left(G\right)}$
where:
\begin{itemize}
\item $S_{1,2}=\liste{s\in S}{\joueur\left(s\right)\in\left\{ 1,2\right\} }$,
\item $\delta\left(G\right)=\liste{d:\etats_{0}\mapsto\etats}{\forall s\in\etats_{0},\ \left(s,d\left(s\right)\right)\in\Actions}$,

\item $s^{\initial}$ is the lower element of $\left(\etats,\Actions\right)$,
\item $s_{d}^{\initial}=\begin{cases}
s^{\initial} & \text{if }s^{\initial}\in S_{1,2}\\
d\left(s^{\initial}\right) & \text{otherwise}
\end{cases}$,
\item $S_{d}=\left\{ s_{\initial}^{d}\right\} \cup
\liste s{\left(s,s'\right)\in\Actions_{d}}\cup\liste{s'}{\left(s,s'\right)\in\Actions_{d}}$,
\item $\Actions_{d}=\bigcup\begin{cases}
\liste{\left(s,s'\right)\in\Actions}{s\in S_{d}\et s'\in S_{1,2}}\\
\liste{\left(s,d\left(s'\right)\right)}{
\left(s,s'\right)\in\Actions \cap \left(S_{d}\times\etats_{0}\right)}
\end{cases}$.
\end{itemize}
\end{definition}

We now know how to transform a stochastic game into a deterministic
game set. It remains to detail the last steps for using the determinization
in the context of reinforcement learning. As said before, during a
reinforcement learning process using the determinization, self-play
matches are performed alternately with learning phases, as usual,
but self-play matches will be performed on determinized games instead
of the stochastic game. Thus, each time a match has to be played,
an element of $D(G)$ is chosen at random and the match is played
on it (as usual using the selected reinforcement learning tools for
deterministic games). The probability of choosing a deterministic
game from $D(G)$ must match the probability that the corresponding
equivalent realization of the stochastic game occurs according to
the stochastic game probabilities. We will provide this probability.
It requires the formalization of two sets: $C_{d}\left(s,s'\right)$
and $J_{d}$. $C_{d}\left(s,s'\right)$ is the set of chance states
of the stochastic game that are a direct transition between the states
$s$ and $s'$, assuming that the outcomes of the random events correspond
to $d$, formally: 
\[
C_{d}\left(s,s'\right)=\liste{c\in\etats_{0}}{\left(s,c\right)\in\Actions\et d\left(c\right)=s'})
\]

$J_{d}$ is the set of chance nodes of the stochastic game which have
been ``jumped'' while playing the associated deterministic game
corresponding to $d$:

$$J_{d}=J_{d}^{\initial}\cup\bigcup_{\left(s,s'\right)\in\Actions_{d}}C_{d}\left(s,s'\right)$$
with
$$J_{d}^{\initial}=\begin{cases}
\left\{ s^{\initial}\right\}  & \text{if }s^{\initial}\in S_{0}\\
\vide & \text{otherwise.}
\end{cases}$$

To choose an element of $D\left(G\right)$, we can choose an element
$d$ of $\delta\left(G\right)$ (the associated game will then be
$\left(S_{d},\Actions_{d},\joueur,\fbin\right)$). We therefore only
need the probability of choosing $d$ in $\delta\left(G\right)$.
The probability of choosing $d$ corresponding to the stochastic game
probabilities is $P\left(d\right)=\prod_{c\in J_d}P\left(c,d\left(c\right)\right)$.

Finally, we have to formalize the modifications of the learning phase
in order to learn knowledge about the stochastic game from the data
generated during the previous determinized game. We recall that classic
reinforcement learning tools for deterministic games provide from
a self-play match the following set of data (which will be used during
the next learning phases): $D=\liste{\left(s,v\left(s\right)\right)}{s\in\Spartiel}$
with $v\left(s\right)$ the learning target value for the state $s$
and $\Spartiel$ the set of states from $\etats_{d}$ encountered
during the last match phase and that was selected for the learning
phase. With the determinization technique, the set $D$ will be increased
with approximate knowledge about chance nodes. More precisely, chance
states ``jumped'' during the performed determinized match and its
associated searches will be added to $D$. The learning target value
of each of these chance states $c$ will be that of its child state
$s'$ corresponding to the chosen determinization $d$ (i.e. $v\left(c\right)$
is $v\left(s'\right)=v\left(d\left(c\right)\right)$). Note the three
following facts. The value of a chance node must be the weighted average
of its child states. The chosen determinization changes at each match
phase according to the probabilities of the stochastic game. The learning
process carried out produces an average effect when several different
values are associated with the same state. From these three facts,
the reinforcement learning processes using the determinization provide
an estimate for the value of chance states. In other words, the
determinization also learns the set of chance states which would have
been encountered if the stochastic game had been played instead of
the deterministic game (and that the players had played the same actions and
the random events had given the outcomes corresponding to $d$). Moreover,
instead of learning the average value for each of these states, the
value of one of its possible samples is learned, that corresponding
to $d$. Formally, $D$ is replaced by the following set: $D_{d}=D\cup D_{d}^{\initial}\cup\liste{\left(c,v\left(s'\right)\right)}{c\in C_{d}\left(s,s'\right)\et s,s'\in \Spartiel}$
with $D_{d}^{\initial}=\begin{cases}
\left\{ \left(s^{\initial},v\left(d\left(s^{\initial}\right)\right)\right)\right\}  & \text{if }s^{\initial}\in S_{0}\\
\vide & \text{otherwise}
\end{cases}$.

\subsubsection{Descent Expectiminimax: Descent with Chance Nodes\label{subsec:Descent-for-Stochastic}}

\begin{table}[t]
\begin{centering}
{\footnotesize{}}%
\begin{tabular}{|c|c|}
\hline 
{\footnotesize{}Symbols} & {\footnotesize{}Definition}\tabularnewline
\hline 
\hline 
{\footnotesize{}$\mathrm{actions}\left(s\right)$} & {\footnotesize{}action set of the state $s$ (for the current player)}\tabularnewline

\hline 
{\footnotesize{}$\mathrm{terminal\left(s\right)}$} & {\footnotesize{}true if $s$ is an end-game state}\tabularnewline
\hline 
{\footnotesize{}$a(s)$} & {\footnotesize{}new state  after playing  action $a$ in the
state $s$}\tabularnewline
\hline 
{\footnotesize{}$\mathrm{time}\left(\right)$} & {\footnotesize{}current time in seconds}\tabularnewline
\hline 


    \multirow{2}{*}{ \footnotesize{}$S$ }
    & {\footnotesize{}states of the partial game tree} \tabularnewline
    & {\footnotesize{}
 (and keys of the transposition table $T$)}
\tabularnewline
\hline 

{\footnotesize{}$T$} & {\footnotesize{}transposition table (contains state labels as $v$)}\tabularnewline

\hline 

{\footnotesize{}$D$} & {\footnotesize{}learning data set}\tabularnewline
\hline 
{\footnotesize{}$\tau$} & {\footnotesize{}search time per action}\tabularnewline

\hline 
{\footnotesize{}$t_{\max}$} & {\footnotesize{}chosen total duration of the learning process }\tabularnewline

\hline 
{\footnotesize{}$v(s)$} & {\footnotesize{}value of state $s$ from the game search}\tabularnewline
\hline 
{\footnotesize{}$v(s,a)$} & {\footnotesize{}value obtained after playing action $a$ in state
$s$}\tabularnewline
\hline

    \multirow{2}{*}{ \footnotesize{}$\hadapt(s)$ }
    & {\footnotesize{}adaptive evaluation function  (of } \tabularnewline
    & {\footnotesize{}
non-terminal states ; first player point of view)}
\tabularnewline
\hline

    \multirow{2}{*}{ \footnotesize{}$\hterminal(s)$ }
    & {\footnotesize{}evaluation of terminal states, e.g. } \tabularnewline
    & {\footnotesize{}
 gain game (first
player point of view)}
\tabularnewline
\hline 

   \multirow{5}{*}{ \footnotesize{}search($s$, $S$, $T$) }
    & {\footnotesize{}a seach algorithm (it extends the game tree  } \tabularnewline
    & {\footnotesize{} from $s$, by adding new states in $S$ and    } \tabularnewline
    & {\footnotesize{}labeling its states by a value $v(s)$, stored in  } \tabularnewline
    & {\footnotesize{}$T$, using $\hadapt$ as evaluation of non-terminal } \tabularnewline
    & {\footnotesize{}leaves  and $\hterminal$ as evaluation of terminal states}
\tabularnewline
\hline 

 \footnotesize{}action\_selection   & {\footnotesize{}decides the action to play in the state $s$ } \\
 {\footnotesize{($s$, $S$, $T$)} }  & {\footnotesize{}
depending on the partial game tree ($S$ and $T$)}
\tabularnewline
\hline 

    \multirow{2}{*}{ \footnotesize{}update($\hadapt,D$) }
    & {\footnotesize{}updates the parameter $\theta$ of $\hadapt$ in order} \tabularnewline
    & {\footnotesize{}
for $\hadapt(s)$ is closer to $v$ for each $(s,v)\in D$}
\tabularnewline
\hline

    \multirow{2}{*}{ \footnotesize{}processing($D$) }
    & {\footnotesize{}various optional data processing: symmetry} \tabularnewline
    & {\footnotesize{}  data augmentation, experience replay, ...}
\tabularnewline
\hline

\end{tabular}{\footnotesize\par}
\par\end{centering}
\caption{Index of symbols\label{tab:Index-of-symbols}}
\end{table}

\begin{figure}
\begin{centering}
\includegraphics[scale=0.5]{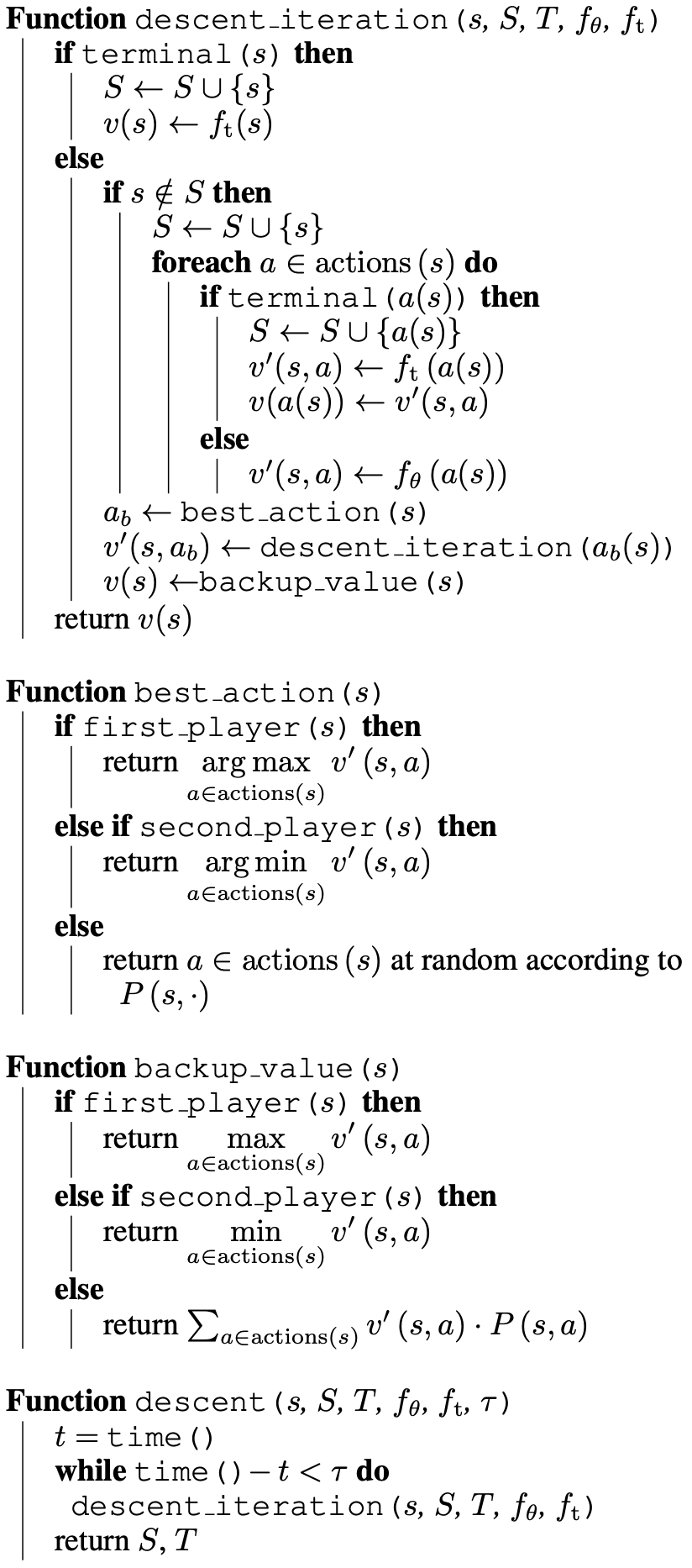}
\caption{\emph{Descent Expectiminimax} tree search algorithm (see Table~\ref{tab:Index-of-symbols}
for the definitions of symbols ; note: $S$ is the set of states which
are non-leaves or terminal and $T=(v,v')$).\label{alg:descente}
$P\left(s,\cdot\right)$ is the probability distribution over $\protect\actions(s)$
given by the game (for all state $s$, we have $\sum_{a\in\protect\actions (s)}P\left(s,a\right)=1$
and for all $a\in\protect\actions(s)$, $0<P\left(s,a\right)\protect\leq1$).
The method first\_player($s$) is true if in the state $s$, it is
the first player to play. The method second\_player($s$) is true
if in the state $s$, it is the second player to play. If first\_player($s$)
and second\_player($s$) are false, it is to the ``chance'' player
to determine (randomly) the next state.}
\par\end{centering}
\end{figure}

\begin{figure}
\begin{centering}
\includegraphics[scale=0.5]{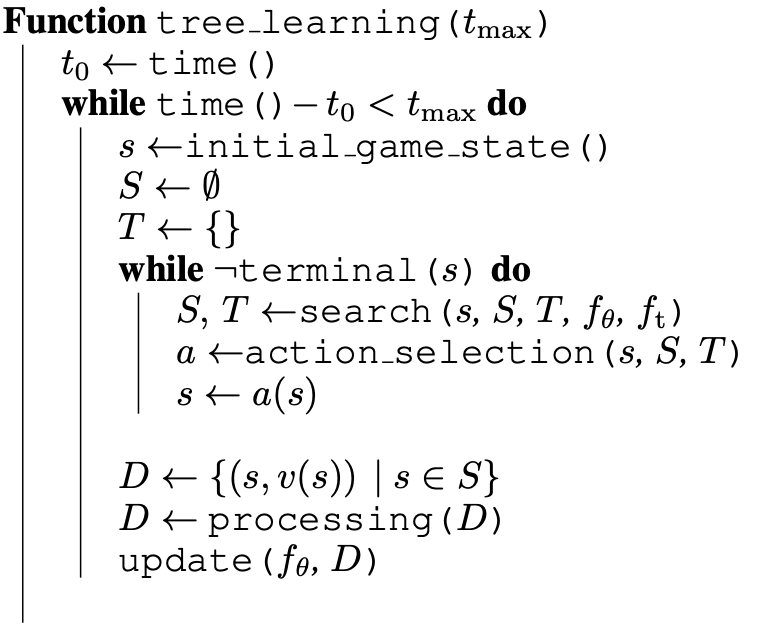}
\caption{Tree learning algorithm (generalized tree bootstrapping) \cite{cohen2020learning} (see Table~\ref{tab:Index-of-symbols}
for the definitions of symbols). For tree learning, $S$ is the set
of states which are non-leaves or terminal.\label{alg:tree-learning}}
\par\end{centering}
\end{figure}

Thus, we generalize the \emph{Descent Minimax} algorithm in the \emph{Descent Expectiminimax} algorithm. As \emph{Descent Minimax}, it performs
deterministic end-game simulations in a best-first order, except that
in added chance states, the next state is chosen at random and that
the values of states are no longer the minimax value but the expectiminimax value (of the partial game tree). 

\emph{Descent Expectiminimax} is described in Algorithm~\ref{alg:descente}. This algorithm 
recursively selects the best child of the current node, which becomes
the new current node. In max states (first player decision states), the best child is the child state of
maximum value. In min state (second player decision states), it is the child of minimum value. In
chance state (states corresponding to chance occurrence), the child state is thus chosen at random according to
the probabilities given by the game. During this recursion, it adds the children of the current
node if they are not in the tree. It performs this recursion from
the root (the current state of the game) until reaching a terminal
node (an end game). It then updates the values of the selected nodes.
In a max state, the updated value is the maximum of the values of the child states.
In a min state, it is the minimum of the values of the child states.
In a chance state, it is the average of the values of the child states
weighted by their probability of occurrence. The \emph{Descent Expectiminimax} algorithm repeats this recursive operation starting from the
root as long as there is some search time left. 

\emph{Descent Expectiminimax}, as \emph{Descent Minimax}, has the advantage of best-first search algorithms,
i.e. it performs long searches, in a best-first order, to determine better actions to play. By learning the values of the game tree (by
using for example tree learning), it also has the advantage of a minimax
search at depth $1$, i.e. it raises the values of the terminal nodes
to the other nodes more quickly. In addition, as for \emph{Descent Minimax},
the states thus generated are closer to the terminal states. Their
values are therefore better approximations than for 
algorithms having shallower searches (classic \emph{Expectiminimax}, MCTS, ...).

To perform a reinforcement
learning process, \emph{Descent Expectiminimax} must be combined with a \emph{learning data selection algorithm}  \cite{cohen2020learning}. As in the perfect-information framework of
Descent, we use generalized tree learning \cite{cohen2020learning}, whose algorithm
is recalled in Figure~\ref{alg:tree-learning}. It consists of
fully learning the partial game tree built during the match, by associating
each game state with its value in the partial game tree, i.e. in this case, its
(partial) expectiminimax value (see in particular the update() method in Algorithm~\ref{alg:tree-learning} and in Table~\ref{tab:Index-of-symbols}).


\subsection{Comparison Experiments\label{subsec:Comparison-Experiments}}

\begin{table}
\begin{centering}
\begin{tabular}{|c|c|c|}
\hline 
 & W.R. & C.R.\tabularnewline
\hline 
\hline 
Descent \& Determinization & $47,3\%$ & $0,7\%$\tabularnewline
\hline 
Expectiminimax & $48,8\%$ & $0,7\%$\tabularnewline
\hline 
Descent Expectiminimax & $53,9\%$ & $0,7\%$\tabularnewline
\hline 
\end{tabular}\caption{Average win rates (W.R.) of different search algorithms for learning
evaluated by the tournament of Section  \ref{subsec:Comparison-between-Descent} and
their associated $95\%$ confidence radius (C.R.) \label{table:tournois}. }
\par\end{centering}
\end{table}

\begin{table*}
\begin{centering}
\begin{tabular}{|c|c|c|c|c|c|c|}
\hline 
Polygames network number & 1 & $2$ & $3$ & C.R. & mean & C.R.\tabularnewline 
\hline 
\hline 
with depth heuristic & $62\%$ & $55,8\%$ & $54,1\%$ & $0,2\%$ & $57,3\%$ & $0,1\%$\tabularnewline
\hline 
without depth heuristic & $61,2\%$ & $54,5\%$ & $51\%$ & $0,2\%$ & $55,6\%$ & $0,1\%$\tabularnewline
\hline 
\end{tabular}\caption{Average win rates 
of neural networks using Expectiminimax for
exploitation and Descent Expectiminimax  for learning (with and without
depth heuristic) againsts Polygames using its three learned neural
networks (and their associated $95\%$ confidence radius (C.R.)) \label{table:polygames}. }
\par\end{centering}
\end{table*}

We now compare the two techniques introduced in the previous section
with each other and with the algorithms of the literature. Before
presenting the experiments and their results, we expose the common
protocol of these experiments as well as the common parameters used,
in particular those of the neural networks and of the reinforcement
learning processes.

\subsubsection{Common Experiment Protocol\label{common_protocol}}

All the experiments performed in this paper consist of performing
zero-knowledge reinforcement learning on the stochastic game \emph{EinStein würfelt nicht!}. The learning process is performed (several times)
for each algorithm to be compared. The duration of each learning process is $30$ days. Each learning process consists
of a sequence of matches where the chosen algorithm plays against itself and learns from the generated data. Each learning
process produces a neural network: an evaluation function containing
the experience of the reinforcement learning process, that can be
used to play matches in combination with a search algorithm. 
See the Appendix for the neural network parameters. 

\paragraph{Evaluation Protocol}
After
the learning processes are ended, each neural network is evaluated
during confrontations on the game \emph{EinStein würfelt nicht!} against all
the other networks (in a Round-robin tournament where each network plays each other network four times (twice as first player and twice as second player). The performance of each search algorithm is then the
average win percentage of each of its corresponding neural networks
(i.e. each neural network using this search algorithm during its learning process). Note that Polygames networks does not participate in this tournament and they will be subject to dedicated additional matches. The search time per action of evaluation matches is 1.5s (on average for Polygames).

\paragraph{Descent framework-based Reinforcement Learning Parameters}

We now expose the parameters of the reinforcement learning processes. Apart from the reinforcement learning processes based on determinization and based on Polygames, we apply the Descent framework without adaptation. For the determinization-based processes, we apply the adapted version of the Descent framework, introduced and explained in the determination sections (Section  \ref{subsec:Approximate-a-Stochastic} and \ref{formalization_determinization}). For Polygames, the standard Polygames architecture is used (details will be provided later in this section).

Here we recall the components of the Descent framework, specifying the parameters used (the learning processes based on determinization uses the same parameters). Its components are a search algorithm for learning, a search algorithm for exploitation (i.e for confrontations: evaluations, tournaments, ...), a \emph{learning data selection algorithm} (which determines what data will be learned and what their learning target will be), an \emph{action selection algorithm} (which selects the next move to play during self-play games in order to manage the exploration-exploitation dilemma), a \emph{reinforcement heuristic} (a terminal state evaluation function which determines the value of the other states), an \emph{update method} (which adapts the adaptive evaluation function to data from previous matches to improve the learned game level), and a \emph{data processing method} (which manages medium-term memory as well as data transformation in order to optimize the learning process) \cite{cohen2020learning}. 
Except for the search algorithms, the default parameters are used. We detail these parameters.
We evaluate the following search algorithms for learning with the Descent framework: \emph{Expectiminimax}, \emph{Descent Expectiminimax}, \emph{Descent} (with determinization). 
The search algorithm for exploitation (for confrontations) is \emph{Expectiminimax}.
We use tree learning as learning data selection algorithm  \cite{cohen2020learning}.
The action selection used is the \emph{ordinal law with simulated annealing on the exploration parameter} \cite{cohen2020learning}. 
The two base reinforcement heuristic are tested:  the \emph{classic gain of the game} (1 for a win, -1 for a loss) and the \emph{depth heuristic} which favor quick wins and slow defeats  \cite{cohen2020learning}. 
The update method is the stochastic gradient descent with ADAM as optimization algorithm \cite{kingma2014adam}.
The data processing method is \emph{smooth experience replay} (its parameters are batch size $B=3000$, duplication factor $\delta=3$, and memory size $\mu=100$)  \cite{cohen2020learning}.

Since, two reinforcement heuristics are tested, for each search algorithm, there are two sets of parameters. Each set of parameters gave rise to four learning processes. Thus, for each search algorithm (except for Polygames), there are 8 processes that have been performed. 


\paragraph{Polygames Reinforcement Learning Parameters \label{pg_params}}

Polygames learning parameters are the default settings. In particular, Monte Carlo with PUCT is used as a search algorithm for learning (with 1600 rollouts per move) and for exploitation. 
3 learning processes were carried out. 


\subsubsection{Our contributions vs Expectiminimax \label{subsec:Comparison-between-Descent}\label{subsec:Comparison-between-Descent-1}}

We start by comparing \emph{Expectiminimax} and the two techniques that we introduced in the
previous section: \emph{Descent Expectiminimax} and the approximation of a stochastic game by a set
of deterministic games on which we apply \emph{Descent Minimax}. 

We apply the common protocol of Section \ref{common_protocol}. Thus, several learning processes are performed based on each of these search algorithms (with and without the depth heuristic) and a tournament is realized based on the associated learned neural networks. The results of this tournament is described in Table \ref{table:tournois}. It is \emph{Descent Expectiminimax} that has the best win rate, then \emph{Expectiminimax}, and finally the worse is \emph{Descent with Determinization}. However, win rates of \emph{Expectiminimax} and \emph{Descent with Determinization} are very close and quite close to \emph{Descent Expectiminimax}, which suggests that \emph{Descent with Determinization} might still have an interest in games with a large stochastic branching factor, as explained in Section \ref{subsec:Approximate-a-Stochastic}.



Note that the use of the depth heuristic significantly improves performance ($44.5\%$ without the depth heuristic and $55.5\%$ with it; $95\%$ confidence radius: $0.5\%$).

\subsubsection{  Descent Expectiminimax vs Polygames\label{subsec:Comparison-between-Descent-2}} 

To end the experiment section, we compare \emph{Descent Expectiminimax}  to Polygames,
using the common experiment protocol (without evaluating the learned neural networks by using a tournament). 

The networks trained from \emph{Descent Expectiminimax} of the previous section have confronted the three networks trained with Polygames. Average win rates of \emph{Descent Expectiminimax} against the three Polygames networks are described in Table \ref{table:polygames}. They are detailed according to whether or not the depth heuristic is used.
Again, the depth heuristic improves performance. And even without the depth heuristic, \emph{Descent Expectiminimax} performs better than Polygames.


\section{Conclusion}

In this article, we have introduced a new search algorithm: \emph{Descent Expectiminimax}, which generalizes Descent Minimax to stochastic games, as well as another learning technique for stochastic games, which makes it possible to use the learning tools of deterministic games. This second technique, 
called Determinization, consists of approximating
stochastic games by sets of deterministic games. Then, we have  evaluated these two techniques on the game \emph{EinStein würfelt nicht!} against state-of-the-art algorithms.

We have shown that in the context of the experiences of this article \emph{Descent Expectiminimax} (with the Descent framework) gave better performance than the Determinization technique using \emph{Descent Minimax} as search algorithm. Although the performance of the Determinization technique is lower, in certain particular contexts, this approximation
technique could perhaps be more effective: when the number of children
of chance nodes is very large (see Section \ref{subsec:Approximate-a-Stochastic}). 

In addition, we have shown that \emph{Descent Expectiminimax} also gives better performance than \emph{Expectiminimax} and than Polygames, the two state-of-the-art algorithms.

Finally, we have also shown that the depth heuristic, a classic tool of the Descent framework, also improves performance.

\section*{Appendix: Neural Networks Parameters\label{neural_parameters}}
Here we detail the parameters of the neural networks.

Except for Polygames, the used adaptive evaluation function (learned by each learning process) is a neural network of the following form: a residual network \cite{he2016deep} with a convolutional layer with 83 filters, followed by 4 residual blocks (2 convolutions per block with 83 filters each), and followed by two fully connected hidden layers (with 425 neurons each). The activation function used is the ReLU \cite{glorot2011deep}.

The network architecture used for Polygames is an adaptations of the architecture being used with Descent Framework-based algorithms, in order to add a policy while keeping an analogous number of weights in the neural network.
Note that,
during each Polygames learning process, several games are performed
in parallel and their evaluations are batched 
to be evaluated by the neural network
in parallel on the GPU.

\bibliographystyle{named}
\bibliography{icml2023}

\begin{thebibliography}{}

\bibitem[\protect\citeauthoryear{Bonnet and Viennot}{2017}]{bonnet2017toward}
Fran{\c{c}}ois Bonnet and Simon Viennot.
\newblock Toward solving “einstein w{\"u}rfelt nicht!”.
\newblock In {\em Advances in Computer Games}, pages 13--25. Springer, 2017.

\bibitem[\protect\citeauthoryear{Cazenave \bgroup \em et al.\egroup
  }{2020}]{cazenave2020polygames}
Tristan Cazenave, Yen-Chi Chen, Guan-Wei Chen, Shi-Yu Chen, Xian-Dong Chiu,
  Julien Dehos, Maria Elsa, Qucheng Gong, Hengyuan Hu, Vasil Khalidov,
  Li~Cheng-Ling, Hsin-I Lin, Yu-Jin Lin, Xavier Martinet, Vegard Mella, Jeremy
  Rapin, Baptiste Roziere, Gabriel Synnaeve, Fabien Teytaud, Olivier Teytaud,
  Shi-Cheng Ye, Yi-Jun Ye, Shi-Jim Yen, and Sergey Zagoruyko.
\newblock Polygames: Improved zero learning.
\newblock {\em ICGA Journal}, 42(4):244--256, December 2020.

\bibitem[\protect\citeauthoryear{Chu \bgroup \em et al.\egroup
  }{2017}]{chu2017agent}
Yeong-Jia~Roger Chu, Yuan-Hao Chen, Chu-Hsuan Hsueh, and I-Chen Wu.
\newblock An agent for einstein w{\"u}rfelt nicht! using n-tuple networks.
\newblock In {\em 2017 Conference on Technologies and Applications of
  Artificial Intelligence (TAAI)}, pages 184--189. IEEE, 2017.

\bibitem[\protect\citeauthoryear{Cohen-Solal and
  Cazenave}{2021}]{cohen2021minimax}
Quentin Cohen-Solal and Tristan Cazenave.
\newblock Minimax strikes back.
\newblock {\em Reinforcement Learning in Games at {AAAI}}, 2021.

\bibitem[\protect\citeauthoryear{Cohen-Solal}{2020}]{cohen2020learning}
Quentin Cohen-Solal.
\newblock Learning to play two-player perfect-information games without
  knowledge.
\newblock {\em arXiv preprint arXiv:2008.01188}, 2020.

\bibitem[\protect\citeauthoryear{Cohen-Solal}{2021}]{cohen2021completeness}
Quentin Cohen-Solal.
\newblock Completeness of unbounded best-first game algorithms.
\newblock {\em arXiv preprint arXiv:2109.09468}, 2021.

\bibitem[\protect\citeauthoryear{Glorot \bgroup \em et al.\egroup
  }{2011}]{glorot2011deep}
Xavier Glorot, Antoine Bordes, and Yoshua Bengio.
\newblock Deep sparse rectifier neural networks.
\newblock In {\em The Fourteenth International Conference on Artificial
  Intelligence and Statistics}, pages 315--323, 2011.

\bibitem[\protect\citeauthoryear{He \bgroup \em et al.\egroup
  }{2016}]{he2016deep}
Kaiming He, Xiangyu Zhang, Shaoqing Ren, and Jian Sun.
\newblock Deep residual learning for image recognition.
\newblock In {\em Conference on Computer Vision and Pattern Recognition}, pages
  770--778, 2016.

\bibitem[\protect\citeauthoryear{Kingma and Ba}{2014}]{kingma2014adam}
Diederik~P Kingma and Jimmy Ba.
\newblock Adam: A method for stochastic optimization.
\newblock {\em arXiv preprint arXiv:1412.6980}, 2014.

\bibitem[\protect\citeauthoryear{Korf and Chickering}{1996}]{korf1996best}
Richard~E Korf and David~Maxwell Chickering.
\newblock Best-first minimax search.
\newblock {\em Artificial intelligence}, 84(1-2):299--337, 1996.

\bibitem[\protect\citeauthoryear{Korf}{1985}]{korf1985depth}
Richard~E Korf.
\newblock Depth-first iterative-deepening: An optimal admissible tree search.
\newblock {\em Artificial Intelligence}, 27(1):97--109, 1985.

\bibitem[\protect\citeauthoryear{Lorentz}{2011}]{lorentz2011mcts}
Richard~J Lorentz.
\newblock An mcts program to play einstein w{\"u}rfelt nicht!
\newblock In {\em Advances in Computer Games}, pages 52--59. Springer, 2011.

\bibitem[\protect\citeauthoryear{Lu \bgroup \em et al.\egroup
  }{2015}]{lu2015einstein}
Junkai Lu, Xiaoyan Wang, Yingyu Li, Yajie Wang, and Tianming Yu.
\newblock Einstein w{\"u}rfelt nicht! strategies research and algorithm
  optimization.
\newblock In {\em The 27th Chinese Control and Decision Conference (2015
  CCDC)}, pages 5818--5821. IEEE, 2015.

\bibitem[\protect\citeauthoryear{Michie}{1966}]{michie1966game}
Donald Michie.
\newblock Game-playing and game-learning automata.
\newblock In {\em Advances in programming and non-numerical computation}, pages
  183--200. Elsevier, 1966.

\bibitem[\protect\citeauthoryear{Piette \bgroup \em et al.\egroup
  }{2020}]{Piette2020Ludii}
{\'E}.~Piette, D.~J. N.~J. Soemers, M.~Stephenson, C.~F. Sironi, M.~H.~M.
  Winands, and C.~Browne.
\newblock Ludii -- the ludemic general game system.
\newblock In G.~De Giacomo, A.~Catala, B.~Dilkina, M.~Milano, S.~Barro,
  A.~Bugarín, and J.~Lang, editors, {\em Proceedings of the 24th European
  Conference on Artificial Intelligence (ECAI 2020)}, volume 325 of {\em
  Frontiers in Artificial Intelligence and Applications}, pages 411--418. IOS
  Press, 2020.

\bibitem[\protect\citeauthoryear{Russell~Stuart and
  Norvig}{2009}]{russell2009artificial}
J~Russell~Stuart and Peter Norvig.
\newblock {\em Artificial intelligence: a modern approach}.
\newblock Prentice Hall, 2009.

\bibitem[\protect\citeauthoryear{Silver \bgroup \em et al.\egroup
  }{2018}]{silver2018general}
David Silver, Thomas Hubert, Julian Schrittwieser, Ioannis Antonoglou, Matthew
  Lai, Arthur Guez, Marc Lanctot, Laurent Sifre, Dharshan Kumaran, Thore
  Graepel, et~al.
\newblock A general reinforcement learning algorithm that masters chess, shogi,
  and go through self-play.
\newblock {\em Science}, 362(6419):1140--1144, 2018.

\bibitem[\protect\citeauthoryear{Slate and Atkin}{1983}]{slate1983chess}
David~J Slate and Lawrence~R Atkin.
\newblock Chess 4.5—the northwestern university chess program.
\newblock In {\em Chess skill in Man and Machine}, pages 82--118. Springer,
  1983.

\bibitem[\protect\citeauthoryear{Turner}{2012}]{turner2012einstein}
Wesley Turner.
\newblock Einstein w{\"u}rfelt nicht--an analysis of endgame play.
\newblock {\em ICGA Journal}, 35(2):94--102, 2012.

\end{thebibliography}

\end{document}